\documentclass{article} 
\usepackage{iclr2021_conference,times}
\usepackage[square,numbers]{natbib}
\usepackage{hyperref}
\usepackage{url}
\usepackage{graphicx}
\usepackage{booktabs}
\usepackage{threeparttable}
\usepackage{amsmath}
\usepackage{amsfonts}
\usepackage{listings}
\usepackage{multirow}

\title{DGL-LifeSci: An Open-Source Toolkit for Deep Learning on Graphs in Life Science}
\author{
Mufei Li \textsuperscript{1},
Jinjing Zhou \textsuperscript{1},
Jiajing Hu \textsuperscript{2},
Wenxuan Fan \textsuperscript{3}, \\
\bf{
Yangkang Zhang \textsuperscript{4},
Yaxin Gu \textsuperscript{3},
George Karypis \textsuperscript{5}
}\\
\textsuperscript{1} AWS Shanghai AI Lab, \textsuperscript{2} King's College London, \\
\textsuperscript{3} East China University of Science and Technology, \\ \textsuperscript{4} Zhejiang University, \textsuperscript{5} AWS AI
}

\iclrfinalcopy 
\begin{document}

\maketitle

\begin{abstract}
Graph neural networks (GNNs) constitute a class of deep learning methods for graph data. They have wide applications in chemistry and biology, such as molecular property prediction, reaction prediction and drug-target interaction prediction. Despite the interest, GNN-based modeling is challenging as it requires graph data pre-processing and modeling in addition to programming and deep learning. Here we present DGL-LifeSci, an open-source package for deep learning on graphs in life science. DGL-LifeSci is a python toolkit based on RDKit, PyTorch and Deep Graph Library (DGL). DGL-LifeSci allows GNN-based modeling on custom datasets for molecular property prediction, reaction prediction and molecule generation. With its command-line interfaces, users can perform modeling without any background in programming and deep learning. We test the command-line interfaces using standard benchmarks MoleculeNet, USPTO, and ZINC. Compared with previous implementations, DGL-LifeSci achieves a speed up by up to 6x. For modeling flexibility, DGL-LifeSci provides well-optimized modules for various stages of the modeling pipeline. In addition, DGL-LifeSci provides pre-trained models for reproducing the test experiment results and applying models without training. The code is distributed under an Apache-2.0 License and is freely accessible at \href{https://github.com/awslabs/dgl-lifesci}{https://github.com/awslabs/dgl-lifesci}.
\end{abstract}

\section{Introduction}

A large amount of the chemical and biological data corresponds to attributed graphs, e.g., molecular graphs, interaction networks and biological pathways. Many of the machine learning (ML) tasks that arise in this domain can be formulated as learning tasks on graphs. For example, molecular property prediction can be formulated as learning a mapping from molecular graphs to real numbers (regression) or discrete values (classification) \cite{NIPS2015_5954}; molecule generation can be formulated as learning a distribution over molecular graphs \cite{pmlr-v80-jin18a}; reaction prediction can be formulated as learning a mapping from one set of graphs (reactants) to another set of graphs (products) \cite{C8SC04228D}. A representation is a vector of a user-defined dimensionality. Graph neural networks (GNNs) combine graph structures and features in representation learning and they have been one of the most popular approaches for learning on graphs \cite{wu2019comprehensive, zhou2019graph}. GNNs have also attracted considerable attention in life science and researchers have applied them to many different tasks \cite{10.1093/bib/bbz042,10.1093/bioinformatics/bty294,pmlr-v80-jin18a,C8SC04228D,STOKES2020688,NIPS2015_5954,pmlr-v70-gilmer17a,shui2020heterogeneous,doi:10.1021/acscentsci.8b00507,NIPS2019_9711}.

Despite significant research, it is often challenging for experts in life science to use GNN-based approaches. To unlock the power of GNNs requires clean interfaces for custom datasets and robust and efficient pipelines. This is because developing GNN pipelines by oneself requires a combined skill set of programming, machine learning, and GNN modeling, which is time-consuming to obtain. This calls for a set of ready-to-run programs, which should make little assumption about users' background.

Prior efforts have greatly lowered the bar for GNN-based modeling in life science, but none of them fully addresses the problem. DeepChem \cite{Ramsundar2019} is a package for deep learning in drug discovery, materials science, quantum chemistry, and biology. While it implements several GNN models, it still requires users to program. Chainer Chemistry \cite{chainerchem} is a package for deep learning in biology and chemistry, based on Chainer \cite{chainer}. It only provides a command-line interface for GNN-based regression on molecules and requires users to write code for other tasks. PiNN \cite{doi:10.1021/acs.jcim.9b00994} implements a GNN variant for predicting potential energy surfaces and physicochemical properties of molecules and materials. It also requires users to program themselves.

Here we present a python toolkit named DGL-LifeSci. It provides high-quality and robust implementations of seven models for molecular property prediction, one model for molecule generation, and one model for chemical reaction prediction. For all these models and tasks, there is an associated command-line script for predictions on custom datasets without writing a single line of code. DGL-LifeSci also provides pre-trained models for all experiments. Compared with previous implementations, it achieves a speedup by up to 6x.

In the following, we first provide a high-level overview of how graph neural networks over molecules work. Then, we discuss the implementation and package features of DGL-LifeSci. After that, we present the results of evaluating DGL-LifeSci interms of robustness and efficiency. Finally, we conclude with a discussion on future work.

\section{Graph Neural Networks over Molecules}

GNNs perform graph-based representation learning by combining information from the topology of a graph and the features associated with its nodes and edges. They iteratively update the representation of a node by aggregating representations of its neighbors. As the number of iterations increases, the nodes gain information from an increasingly larger local subgraph.

\begin{figure}
  \centering
  \includegraphics[width=0.75\linewidth]{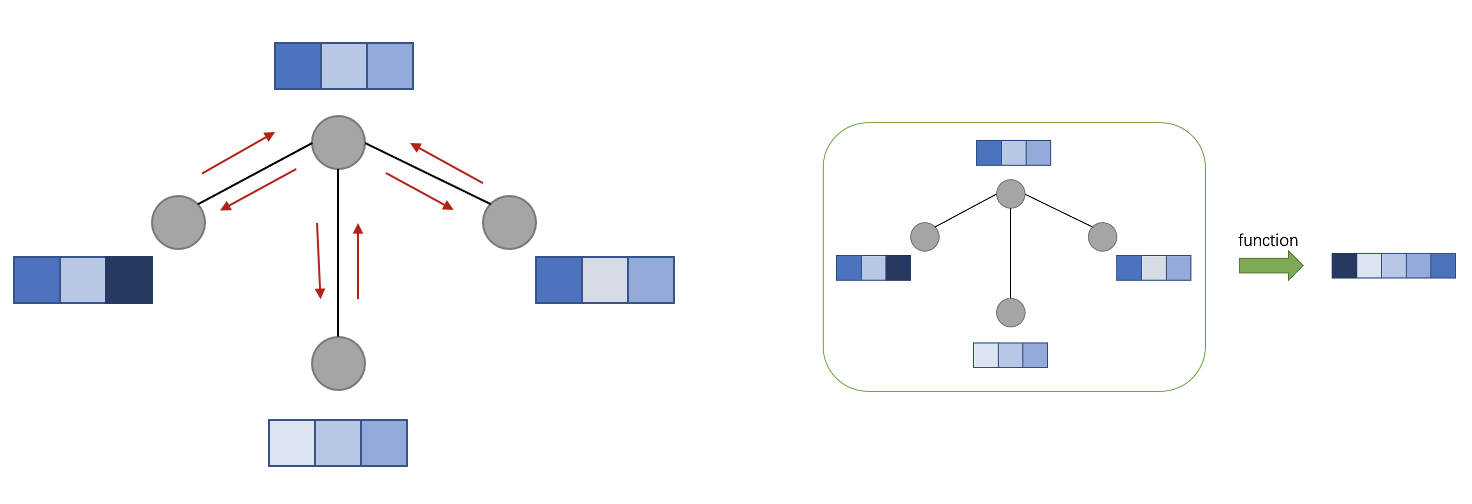}
  \caption{Illustration of the message passing phase (left) and readout phase (right).}
  \label{fgr:gnn}
\end{figure}

When applying GNNs to molecules as in molecular property prediction, there are two phases -- a message passing phase and a readout phase. Figure~\ref{fgr:gnn} is an illustration of them.

\textbf{Message Passing Phase.} The message passing phase updates node representations simultaneously across the entire graph and consists of multiple rounds of message passing. In a round of message passing, the representation of a node is updated by applying learnable functions to its original representation, the representations of its adjacent nodes and the representations of its incident edges. The operation is similar to gathering messages from adjacent nodes. By performing $k$ rounds of message passing, we can aggregate information from all the nodes/edges that are within $k$ hops from each node.
 
\textbf{Readout Phase.} The readout phase computes a representation for the entire graph. This representation is computed by applying a potentially learnable function to the representations of all the nodes in the graph, e.g., summation over them. Once we obtain graph representations, we can pass them to a multilayer perceptron (MLP) for final prediction.

\section{Package Features}

DGL-LifeSci contains four components: (i) a set of ready-to-run scripts for training and inference; (ii) programming APIs for allowing researchers to develop their own custom pipelines and models; (iii) a set of pre-trained models that can either be fine-tuned or directly used to perform inference; (iv) a set of built-in datasets for quick experimentation. 

It provides models that can be used to solve three tasks. The first task is molecular property prediction or quantitative structure activity relationship (QSAR) prediction. This can be formulated as a regression or classification task for single molecules. The second task is molecule generation. The third task is chemical reaction prediction.

\textbf{Usage.} DGL-LifeSci provides command-line scripts for each task. They are responsible for invoking the modeling pipeline, which handles model training and model evaluation. Users need to prepare their data in a standard format. They can then use the command-line interface by specifying the path to the data file along with some additional arguments. For example, below is the command-line interface for regression and classification problems in molecular property prediction. Users need to prepare molecules in the form of SMILES strings with their properties to predict in a CSV file. 
\begin{lstlisting}[language=bash]
python regression_train.py -c file -sc header -mo model
python classification_train.py -c file -sc header -mo model
\end{lstlisting}
DGL-LifeSci also provides an optional support for hyperparameter search other than using the default ones. It uses Bayesian optimization based on hyperopt \cite{Bergstra12} for hyperparameter search.

\section{Implementation}

\textbf{Dependencies.} DGL-LifeSci is developed using PyTorch \cite{NIPS2019_9015} and Deep Graph Library (DGL) \cite{wang2020dgl}. PyTorch is a general-purpose deep learning framework and DGL is a high-performant GNN library. In addition, it uses RDKit \cite{rdkit} for utilities related to cheminformatics.

\begin{figure}
  \centering
  \includegraphics[width=0.75\linewidth]{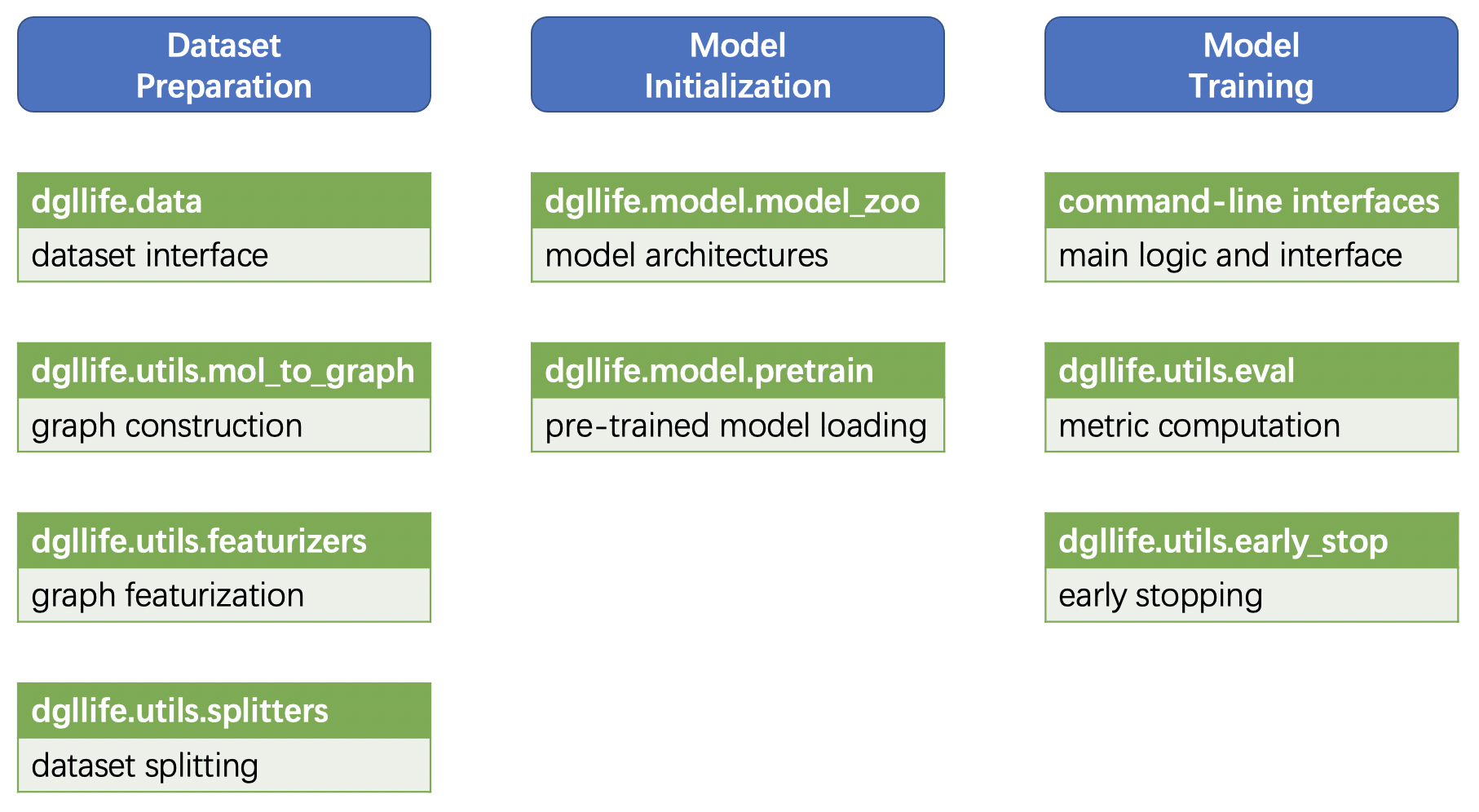}
  \caption{An overview of modules in DGL-LifeSci and their usage.}
  \label{fgr:design}
\end{figure}

\textbf{Modeling Pipeline and Modules.} A general GNN-based modeling pipeline consists of three stages: dataset preparation, model initialization and model training. The dataset preparation stage involves data loading, graph construction, representation initialization for nodes and edges (graph featurization) and dataset interface construction. The model training stage involves model update, metric computation and early stopping. As presented in figure \ref{fgr:design}, DGL-LifeSci is modularized for these various stages and stage components so as to cater to the need of different uses. While DGL-LifeSci allows users to perform GNN-based modeling without programming, advanced users can also adapt these modules for their own development.

\textbf{Dataset Preparation.} DGL-LifeSci provides dataset interfaces for supporting both built-in datasets and custom datasets. The interfaces are responsible for loading raw data files and invoking graph construction and featurization. 

Graph construction and featurization are two important steps for GNN-specific data preparation. DGL-LifeSci provides built-in support for constructing three kinds of graphs for molecules -- molecular graphs, distance-based graphs, and complete graphs. In all these graphs, each node corresponds to an atom in a molecule. In a molecular graph, the edges correspond to chemical bonds in the molecule. The construction of a distance-based graph requires a molecule conformation and there is an edge between a pair of atoms if the distance between them is within a cutoff distance. In a complete graph, every pair of atoms is connected. For graph featurization, DGL-LifeSci allows initializing various node and edge features from atom and bond descriptors. Table \ref{tbl:feature} gives an overview of them.

\begin{table}
\begin{center}
\footnotesize
\caption{Descriptors for feature initialization.
\label{tbl:feature}}
\begin{threeparttable}
\begin{tabular}{lc}
\toprule
Descriptors & Possible values \\ 
\midrule
Atom type   & C, N, O, etc  \\
Atom degree excluding hydrogen atoms & Non-negative integers \\ 
Atom degree including hydrogen atoms & Non-negative integers \\
Atom explicit valence & Non-negative integers \\
Atom implicit valence & Non-negative integers \\
Atom hybridization & S, SP, SP2, SP3, SP3D, SP3D2 \\
Total number of hydrogen atoms attached & Non-negative integers \\
Atom formal charge & Integers \\
Number of radical electrons of an atom & Non-negative integers \\ 
Whether an atom is aromatic & 1 (True), 0 (False) \\
Whether an atom is in a ring & 1 (True), 0 (False) \\
Atom chiral tag & CW, CCW, unspecified, other \\
Atom chirality type & R, S \\
Atom mass & Non-negative real numbers \\
Whether an atom is chiral center & 1 (True), 0 (False) \\
\midrule
Bond type & single, double, triple, aromatic \\
Whether a bond is conjugated & 1 (True), 0 (False) \\
Whether a bond is in a ring & 1 (True), 0 (False) \\
Stereo configuration of a bond & none, any, OZ, OE, CIS, TRANS \\
Direction of a bond & none, end-up-right, end-down-right \\
\bottomrule
\end{tabular}
\footnotesize
For non-numeric discrete-valued descriptors, one-hot encoding is used in featurization. For numeric discrete-valued descriptors, either raw number or one-hot encoding can be used in featurization.
\end{threeparttable}
\normalsize
\end{center}
\end{table}
Users can split the dataset into training/validation/test subsets or do so for $k$-fold cross validation. DGL-LifeSci provides built-in support for random split, scaffold split, weight split and stratified split \cite{Wu18}. The random split performs a pure random split of a dataset. The scaffold split separates structurally different molecules into different subsets based on their Bemis-Murcko scaffolds \cite{Bemis96}. The weight split sorts molecules based on their weight and then splits them in order. The stratified split sorts molecules based on their label and ensures that each subset contains nearly the full range of provided labels.

\textbf{Models Included.} Table \ref{tbl:model} lists the models implemented. GCN and GAT are two popular GNNs initially developed for node classification. We extend them for graph regression/classification with a readout function and an MLP. NF and Weave are among the earliest models that extend rule-based molecular fingerprints with graph neural networks. MPNN unifies multiple GNNs for quantum chemistry. AttentiveFP extends GAT with gated recurrent units \cite{cho-etal-2014-learning}. 

One difficulty in developing learning-based approaches for molecular property prediction is the gap between an extremely large chemical space and extremely limited labels for molecular properties. It is estimated that the number of drug-like molecules is between $10^{23}$ and $10^{60}$ while most datasets have less than tens of thousands of molecules in MoleculeNet \cite{Polishchuk13,Hagan18,Reymond15,Dobson04,Wu18}. Hu et al. \cite{Hu20} propose to approach this problem by utilizing millions of unlabeled molecules in pre-training the weights of a GIN model for general molecule representations. One can then fine-tune the model weights for predicting particular properties. We include four pre-trained models from their work in DGL-LifeSci. The models were pre-trained with a same strategy for supervised learning and a different strategy for self-supervised learning. We distinguish the models by the associated strategy for self-supervised learning, which are context prediction, deep graph infomax, edge prediction and attribute masking.

JTVAE is an autoencoder that utilizes both a junction tree and a molecular graph for the intermediate representation of a molecule. WLN is a two-stage model for chemical reaction prediction. It first identifies potential bond changes and then enumerates and ranks candidate products.

\begin{table}[t]
\begin{center}
\footnotesize
\caption{Models Implemented.
\label{tbl:model}}
\begin{threeparttable}
\begin{tabular}{lc}
\toprule
Task & Model \\ 
\midrule
Molecular property prediction & GCN \cite{Kipf17}, GAT \cite{Velivckovic18}, NF \cite{NIPS2015_5954}, Weave \cite{Weave}, MPNN \cite{MPNN}, AttentiveFP \cite{Xiong19} \\
                              & GIN + context prediction/deep graph infomax/\\
                              & edge prediction/attribute masking \cite{Hu20} \\
Molecule generation & JTVAE \cite{pmlr-v80-jin18a} \\
Reaction prediction & WLN \cite{C8SC04228D} \\
\bottomrule
\end{tabular}
\end{threeparttable}
\normalsize
\end{center}
\end{table}

\section{Modeling Performance}

\textbf{Molecular Property Prediction.} We test against six binary classification datasets in MoleculeNet and evaluate the model performance by ROC-AUC averaged over all tasks \cite{Wu18}. To evaluate the model performance on unseen structures, we employ the scaffold split and use respectively 80\%, 10\% and 10\% of the dataset for training, validation, and test. We train six models (NF, GCN, GAT, Weave, MPNN, AttentiveFP) from scratch using the featurization proposed in DeepChem, which is described in table \ref{tbl:mol_feat}. GCN and GAT take initial node features only and they do not take initial edge features. We also fine-tuned the four pre-trained GIN models. For a non-GNN baseline model, we train an MLP taking Extended-Connectivity Fingerprints (ECFPs).

For all the settings, we perform a hyperparameter search for 32 trials. Within each trial, we train a randomly initialized model and perform an early stopping if the validation performance no longer improves for 30 epochs. Finally, we evaluate the model achieving the best validation performance across all epochs and trials on the test set. Table \ref{tbl:molnet_roc_auc} presents the summary of the test performance. For a reference, we also include the fine-tuning performance reported previously\cite{Hu20}.

\begin{table}
\begin{center}
\footnotesize
\caption{Descriptors Considered in DeepChem Featurization.
\label{tbl:mol_feat}}
\begin{threeparttable}
\begin{tabular}{lc}
\toprule
Descriptors & Possible values \\ 
\midrule
Atom type (one-hot encoding) & C, N, O, S, F, Si, P, Cl, Br, Mg, Na, Ca, \\ 
                             & Fe, As, Al, I, B, V  K, Tl, Yb, Sb, Sn, \\
                             & Ag, Pd, Co, Se, Ti, Zn, H, Li, Ge, Cu, \\
                             & Au, Ni, Cd, In, Mn, Zr, Cr, Pt, Hg, Pb  \\
Atom degree excluding hydrogen atoms (one-hot encoding) & 0 - 10 \\ 
Atom implicit valence (one-hot encoding) & 0 - 6 \\
Atom formal charge & Integers \\
Number of radical electrons of an atom & Non-negative integers \\ 
Whether an atom is aromatic & 1 (True), 0 (False) \\
Atom hybridization (one-hot encoding) & SP, SP2, SP3, SP3D, SP3D2 \\
Total number of hydrogen atoms attached (one-hot encoding) & 0 - 4 \\
\midrule
Bond type (one-hot encoding) & single, double, triple, aromatic \\
Whether a bond is conjugated & 1 (True), 0 (False) \\
Whether a bond is in a ring  & 1 (True), 0 (False) \\
Stereo configuration of a bond (one-hot encoding) & none, any, OZ, OE, CIS, TRANS \\
\bottomrule
\end{tabular}
\end{threeparttable}
\normalsize
\end{center}
\end{table}

\begin{table}
\begin{center}
\footnotesize
\caption{Test ROC-AUC on 6 Datasets from MoleculeNet.
\label{tbl:molnet_roc_auc}}
\begin{threeparttable}
\begin{tabular}{lcccccc}
\toprule
Model & BBBP & Tox21 & ToxCast & SIDER & HIV & BACE \\
\midrule
\multicolumn{7}{c}{Models trained from scratch} \\
\midrule
GCN                      & 0.63 & 0.77 & 0.62 & 0.58 & 0.76 & 0.84\\
GAT                      & 0.68 & 0.71 & 0.64 & 0.52 & 0.76 & 0.84\\
NF                       & 0.66 & 0.75 & 0.60 & 0.53 & 0.74 & 0.80\\
Weave                    & 0.67 & 0.56 & 0.62 & 0.58 & 0.73 & 0.79\\
MPNN                     & 0.65 & 0.70 & 0.59 & 0.54 & 0.74 & 0.85\\
AttentiveFP              & 0.71 & 0.70 & 0.57 & 0.53 & 0.75 & 0.73\\
\midrule
\multicolumn{7}{c}{Non-GNN baseline} \\
\midrule
MLP + ECFP               & 0.67 & 0.70 & 0.58 & 0.63 & 0.76 & 0.80\\
\midrule
\multicolumn{7}{c}{Pre-trained models fine-tuned} \\
\midrule
GIN + context prediction & 0.63 & 0.75 & 0.64 & 0.61 & 0.77 & \textbf{0.86}\\
GIN + deep graph infomax & \textbf{0.72} & 0.78 & 0.59 & 0.63 & 0.76 & 0.71\\
GIN + edge prediction    & 0.70 & \textbf{0.80} & 0.59 & \textbf{0.66} & 0.72 & \textbf{0.86}\\
GIN + attribute masking  & \textbf{0.72} & 0.75 & 0.58 & 0.58 & 0.75 & 0.74\\
\midrule
\multicolumn{7}{c}{Previously reported results} \\
\midrule
GIN + context prediction & 0.69 & 0.78 & 0.66 & 0.63 & \textbf{0.80} & 0.85\\
GIN + deep graph infomax & 0.68 & 0.78 & 0.65 & 0.61 & 0.78 & 0.80\\
GIN + edge prediction    & 0.67 & 0.78 & \textbf{0.67} & 0.63 & 0.78 & 0.79\\
GIN + attribute masking  & 0.67 & 0.78 & 0.65 & 0.64 & 0.77 & 0.80\\
\bottomrule
\end{tabular}
\end{threeparttable}
\normalsize
\end{center}
\end{table}

\textbf{Reaction Prediction.} We test WLN against USPTO \cite{USPTO} dataset following the setting in the original work \cite{C8SC04228D}. WLN is a two-stage model for reaction prediction. The first stage identifies candidate reaction centers, i.e. pairs of atoms that lose or form a bond in the reaction. The second stage enumerates candidate products from the candidate reaction centers and ranks them. We achieve comparable performance for both stages as in table \ref{tbl:reaction}.

\begin{table}[t]
\caption{Test Top-k Accuracy (\%) of WLN on USPTO.} 
\label{tbl:reaction}
\footnotesize
\centering 
\begin{tabular}{ccccccccc} %
\toprule
\multirow{2}{*}{
\parbox[c]{.2\linewidth}{\centering Implementations}}
  & \multicolumn{3}{c}{Reaction center prediction} &&
\multicolumn{4}{c}{Candidate ranking} \\ 
\cmidrule{2-4} \cmidrule{6-9}

 & {\centering Top 6} & {Top 8} & {Top 10} && {Top 1} & {Top 2} & {Top 3} & {Top 5 } \\
\midrule
Original    & 89.8 & 92.0 & 93.3 && 85.6 & 90.5 & 92.8 & 93.4 \\
DGL-LifeSci & 91.2 & 93.8 & 95.0 && 85.6 & 90.0 & 91.7 & 92.9 \\
\bottomrule
\end{tabular}
\end{table}

\textbf{Molecule Generation.} We test JTVAE against a ZINC \cite{ZINC} subset for reconstructing input molecules \cite{pmlr-v80-jin18a}. We achieve an accuracy of 76.4\% while the authors' released code achieve an accuracy of 74.4\%.

\section{Training Speed}

We compare the modeling efficiency of DGL-LifeSci against previous implementations, including original implementations and DeepChem. All experiments record the averaged training time of one epoch. The testbed is one AWS EC2 p3.2xlarge instance (one NVidia V100 GPU with 16GB GPU RAM and 8 VCPUs).

\begin{table}[t]
\begin{center}
\footnotesize
\caption{Epoch Training Time in Seconds.
\label{tbl:efficiency}}
\begin{threeparttable}
\begin{tabular}{lcccc}
\toprule
Experiment & Dataset & Previous implementation & DGL-LifeSci & Speedup \\
\midrule
\multicolumn{5}{c}{Molecular property prediction} \\
\midrule
NF & HIV & 5.8 (DeepChem 2.3.0) & 2.5 & 2.3x \\
AttentiveFP & Aromaticity \cite{Xiong19} & 6.0 & 1.0 & 6.0x \\
\midrule
\multicolumn{5}{c}{Reaction prediction} \\
\midrule
WLN for reaction center prediction & USPTO & 11657 & 2315 & 5.0x \\
\midrule
\multicolumn{5}{c}{Molecule generation} \\
\midrule
JTVAE & ZINC subset & 44666 & 44843 & 1.0x \\ 
\bottomrule
\end{tabular}
\end{threeparttable}
\normalsize
\end{center}
\end{table}

\section{Conclusions}

Here, we present DGL-LifeSci, an open-source Python toolkit for deep learning on graphs in life science. In the current version of DGL-LifeSci, we support GNN-based modeling for molecular property prediction, reaction prediction and molecule generation.

With command-line interfaces, users can perform efficient modeling on custom datasets without programming a single line of code. Advanced users can also adapt highly modularized building blocks for their own development.

In the current implementations of DGL-LifeSci, the primary focus is on small molecules. In the future, we aim to extend the support to other graphs in life science like proteins and biological networks. This will open up a much richer set of tasks in life science.

\section{Data and Software Availability}

The datasets and models are publicly available at \href{https://github.com/awslabs/dgl-lifesci}{https://github.com/awslabs/dgl-lifesci}. The scripts for reproducing the experiments are available in the following examples.
\begin{itemize}
    \item Molecular property prediction: examples/property\underline{ }prediction/moleculenet
    \item Reaction prediction: examples/reaction\underline{ }prediction/rexgen\underline{ }direct
    \item Molecule generation: examples/generative\underline{ }models/jtvae
\end{itemize}

\bibliography{main}

\end{document}